\documentclass{article}

\usepackage{arxiv}

\usepackage[utf8]{inputenc}    
\usepackage[T1]{fontenc}       
\usepackage{url}               
\usepackage{booktabs}          
\usepackage{multirow}          
\usepackage{amsfonts}          
\usepackage{amsmath}           
\usepackage{amssymb}           
\usepackage{nicefrac}          
\usepackage{microtype}         
\usepackage{graphicx}
\usepackage[round,semicolon,authoryear]{natbib}  
\usepackage{xcolor}
\usepackage{xspace}            
\usepackage{algorithm}         
\usepackage{algpseudocode}     
\usepackage{caption}           
\usepackage[colorlinks=true,
            linkcolor=black,
            citecolor=blue,
            urlcolor=blue,
            pdfborder={0 0 0}]{hyperref}
\graphicspath{{./images/}}

\newcommand{\GRPO}{GRPO\xspace}
\newcommand{\DAPO}{DAPO\xspace}
\newcommand{\DGPO}{DGPO\xspace}
\newcommand{\AKL}{AKL\xspace}
\newcommand{\GCS}{GCS\xspace}
\newcommand{\accbar}{\bar{\mathrm{acc}}}
\newcommand{\KLdiv}{D_{\mathrm{KL}}}
\newcommand{\Ahat}{\hat{A}}

\newcommand{\posdelta}[1]{\textcolor{green!50!black}{#1}}
\newcommand{\negdelta}[1]{\textcolor{red!75!black}{#1}}

\title{FG-ExPO: Frontier-Guided Exploration-Prioritized Policy Optimization via\\
Adaptive KL and Gaussian Curriculum}

\author{%
  Mingxiong Lin,\quad Zhangquan Gong,\quad Maowen Tang,\quad Qian Li,\quad Chuangchuang Wang, \\[2pt]
  \bfseries Jian Ma,\quad Sutian Huang,\quad Kai Tang\textsuperscript{$\dagger$},\quad Haonan Lu\textsuperscript{$\dagger$} \\[6pt]
  \normalfont OPPO AI Center \\[4pt]
  \normalfont\texttt{\{linmax1111,\,lixiaoqian0208\}@gmail.com} \\[2pt]
  \normalfont\texttt{m.w.tang@i4ai.org} \\[2pt]
  \normalfont\texttt{\{gongzhangquan,\,wangchuangchuang,\,majian2,\,sutianhuang,\,tangkai.a,\,luhaonan\}@oppo.com} \\[6pt]
  \normalfont\small\textsuperscript{$\dagger$}\,Corresponding authors.
}
\date{}  

\begin{document}
\maketitle

\begin{abstract}
Reinforcement Learning with Verifiable Rewards~(RLVR) has become the
standard recipe for LLM mathematical reasoning, with Group Relative
Policy Optimization~(\GRPO) as the dominant algorithm.
We identify two overlooked inefficiencies in \GRPO:
\emph{(i)} a fixed KL coefficient over-constrains exploration when the
model most needs to deviate from the reference policy; and
\emph{(ii)} uniform question sampling neglects that
moderate-difficulty problems yield the richest gradient signal.
We propose \textbf{FG-ExPO}
(\textbf{F}rontier-\textbf{G}uided
\textbf{Ex}ploration-\textbf{P}rioritized Policy \textbf{O}ptimization),
which couples two lightweight components:
\textbf{Accuracy-Conditioned KL Scaling~(\AKL)} modulates the KL
penalty via a smooth nonlinear function of the batch's mean
accuracy---relaxing it when the model struggles and tightening it when
the model succeeds; and
\textbf{Gaussian Curriculum Sampling~(\GCS)} weights questions by a
Gaussian centered at moderate accuracy~($p\!\approx\!0.5$),
concentrating learning on the model's frontier.
Experiments on
DeepSeek-R1-Distill-Qwen-1.5B~\citep{deepseekai2025r1distill} and
Qwen3-8B-Base~\citep{yang2025qwen3} across six competitive
benchmarks show
that FG-ExPO consistently outperforms \GRPO, achieving a $\mathbf{+13.34}$
absolute improvement on AIME~2025 pass@32 (63.33\%~$\to$~76.67\%) and a
$+$2.66 average pass@32 gain on the 8B model.
The disproportionately larger gains on pass@32 over pass@1 confirm that
FG-ExPO expands the model's effective exploration space within a fixed
inference budget.
\end{abstract}

\section{Introduction}
\label{sec:intro}

As large language models~(LLMs) advance, demand for them to produce not
only correct answers but also reliably elicited multi-step reasoning has
grown rapidly, particularly on competition-level mathematics,
programming, and scientific
tasks~\citep{jaech2024openai,guo2025deepseek}.
Reinforcement learning with verifiable rewards~(RLVR) has emerged as the
de facto post-training pipeline that drives this
progress~\citep{shao2024deepseekmath,wen2025rlvr}: by replacing learned
reward models with rule-based correctness signals, RLVR avoids reward
hacking and scales reliably to long chain-of-thought reasoning behaviors
such as self-verification and reflection.
Meeting these stringent demands within a single training run, however,
imposes strong requirements on how the policy is regularized and what
data it is trained on.

Group Relative Policy Optimization~(\GRPO; \citealt{shao2024deepseekmath})
has become the dominant algorithm for RLVR, eliminating the value critic
of PPO~\citep{schulman2017ppo} via group-relative advantage estimation.
Building on \GRPO, recent
work~\citep{yu2025dapo,chu2025gpg,liu2025drgrpo,dai2026mathforge} has
focused largely on advantage normalization, clipping schedules, loss
reduction, and reward design.
What has received little attention, however, is the interaction between
two seemingly innocuous choices that \GRPO inherits unchanged: a
\emph{fixed KL divergence coefficient} that regularizes the policy by
the same amount throughout training, and a \emph{uniform sampling
distribution} over training questions that ignores how each question's
difficulty evolves as the model improves.
Both choices are widely adopted---and as we show, both are systematically
suboptimal.

In this paper, we revisit these two design choices and identify a single
unifying failure mode: \GRPO fails to allocate exploration budget where
it matters most.
For the KL coefficient, we observe that the regularization strength
required for healthy learning depends sharply on the model's current
competence: when the model fails most problems in a batch, the
reference-model anchor must be \emph{relaxed} so the policy can deviate
toward new reasoning patterns; when the model already succeeds, the
anchor must be \emph{tightened} to suppress overfitting.
A constant $\beta$ collapses these two regimes into one, producing an
exploration--stability trade-off that is chronically miscalibrated
throughout training.
For uniform question sampling, we observe an analogous waste: questions
the model almost always solves yield near-zero advantages, and
questions it almost never solves yield essentially no positive
reinforcement signal---only intermediate-difficulty questions
($p\approx 0.5$, where $p$ is the empirical pass rate) lie on the
model's \emph{learning frontier} and produce informative gradients.
Together, these two miscalibrations cause the policy to spend its
gradient budget far from where it could most improve.
Existing remedies address only fragments of this picture: \DAPO removes
KL entirely~\citep{yu2025dapo}, sacrificing stability for exploration,
and applies binary hard filtering to drop $p\!=\!0$ and $p\!=\!1$
batches without weighting the spectrum in between;
MathForge's \DGPO~\citep{dai2026mathforge} monotonically upweights hard
problems but still wastes capacity on $p\!\approx\!0$ batches that
yield no positive signal.

To overcome these challenges, we propose \textbf{FG-ExPO}
(\textbf{F}rontier-\textbf{G}uided
\textbf{Ex}ploration-\textbf{P}rioritized Policy \textbf{O}ptimization),
a unified extension of \GRPO that addresses both miscalibrations with
two complementary, hyperparameter-light components.
The first component, \textbf{Accuracy-Conditioned KL Scaling~(\AKL)},
makes the KL coefficient a smooth, monotone function of the batch's
mean accuracy, dynamically reducing the effective penalty when the
model is failing (encouraging exploration) and increasing it when the
model is succeeding (preserving stability).
The second component, \textbf{Gaussian Curriculum Sampling~(\GCS)},
re-weights training questions by a Gaussian density centered at
$p\!=\!0.5$, giving frontier-difficulty questions the highest
probability of being trained on and smoothly down-weighting both
already-mastered ($p\!\approx\!1$) and currently-intractable
($p\!\approx\!0$) problems.
Together, \AKL controls how aggressively the policy may move per
update, while \GCS controls where in the data distribution that
movement is spent---producing a coordinated allocation of exploration
budget that neither component achieves alone.

We compare FG-ExPO against \GRPO on the DAPO-17K
dataset~\citep{yu2025dapo} using two base models of distinct families
and scales---DeepSeek-R1-Distill-Qwen-1.5B~\citep{deepseekai2025r1distill}
and Qwen3-8B-Base~\citep{yang2025qwen3}---and six competitive
mathematical reasoning benchmarks: AIME~2024, AIME~2025, MATH-500,
Minerva, OlympiadBench, and AMC.
FG-ExPO consistently outperforms \GRPO across both scales and all six
benchmarks.
For example, on the 8B model FG-ExPO achieves a $\mathbf{+13.34}$ absolute
improvement on AIME~2025 pass@32 (63.33\%~$\to$~76.67\%), $+$3.33
points on AIME~2025 pass@1, and a $+$2.66 average pass@32 gain across
all six benchmarks; on the 1.5B model FG-ExPO yields a $+$2.16 average
pass@32 gain, with $+$10.00 absolute improvement on AIME~2024 pass@32.
The substantially larger gains on pass@32 than on pass@1 directly
corroborate our central claim: by reallocating exploration budget toward
where it matters most, FG-ExPO does not merely refine the model's most
likely solution---it expands the diversity of correct reasoning paths
the model can discover within a fixed inference budget.

\medskip\noindent
Our main contributions are as follows:
\begin{itemize}
  \item \textbf{Accuracy-Conditioned KL Scaling~(\AKL).}
    A parameter-free adaptive KL mechanism that conditions the
    regularization strength on the batch's mean accuracy via a smooth
    nonlinear scaling function, automatically tightening or relaxing
    the reference-model anchor in proportion to the model's current
    competence.

  \item \textbf{Gaussian Curriculum Sampling~(\GCS).}
    A smooth Gaussian-shaped weighting scheme that concentrates training
    on frontier-difficulty questions ($p\!\approx\!0.5$), generalizing
    both binary hard filtering and monotone difficulty upweighting in a
    single continuous formulation.

  \item \textbf{Extensive empirical validation.}
    Across two model scales and six competitive benchmarks, FG-ExPO
    consistently outperforms \GRPO, achieving up to $+$13.34 absolute
    improvement on AIME~2025 pass@32 and $+$2.66 average pass@32 gain
    on the 8B model.
\end{itemize}

\section{Related Work}
\label{sec:related}

\paragraph{RLVR for LLM mathematical reasoning.}
Reinforcement learning with verifiable rewards~(RLVR) has become the
dominant paradigm for eliciting long chain-of-thought reasoning in
large language models~\citep{shao2024deepseekmath,wen2025rlvr,guo2025deepseek,jaech2024openai}.
Building on PPO~\citep{schulman2017ppo}, Group Relative Policy
Optimization~(\GRPO; \citealt{shao2024deepseekmath}) replaces the
value critic with a group-relative advantage estimator and has become
the de facto backbone for math RLVR pipelines.
A growing line of follow-up work largely focuses on the
\emph{advantage and loss side} of \GRPO: \DAPO removes the KL term and
reshapes the clipping schedule~\citep{yu2025dapo}; GPG revises the
group-policy gradient~\citep{chu2025gpg}; DR.GRPO refines advantage
normalization to mitigate length and difficulty
biases~\citep{liu2025drgrpo}; and
MathForge introduces problem-level reweighting via
\DGPO~\citep{dai2026mathforge}.
What these methods almost universally inherit unchanged from \GRPO are
two seemingly secondary choices---a \emph{fixed KL coefficient} and a
\emph{uniform question sampling distribution}.
Our work targets exactly this overlooked pair, treating the KL strength
and the sampling distribution as first-class objects of optimization
within the same RLVR framework.

\paragraph{KL regularization in policy optimization.}
Constraining a policy near a reference distribution via a KL term is a
standard ingredient of policy optimization, dating back to PPO's
adaptive KL targeting~\citep{schulman2017ppo} and re-emerging in
\GRPO~\citep{shao2024deepseekmath} as a fixed coefficient~$\beta$ on
$\KLdiv(\pi_\theta\|\pi_{\mathrm{ref}})$.
Recent RLVR variants treat KL primarily as a stability knob: \DAPO
\emph{removes} the KL term entirely~\citep{yu2025dapo}, trading
reference-anchored stability for unconstrained exploration; other works
inherit a single global $\beta$ throughout training.
This binary view---KL on at full strength, or KL off---ignores the fact
that the regularization strength required for healthy learning depends
on the model's \emph{current competence}: the policy should be allowed
to deviate further when it is failing on a batch and held closer to
$\pi_{\mathrm{ref}}$ when it is already succeeding.
Our \AKL component instead conditions the effective KL coefficient on
the batch's mean accuracy through a smooth nonlinear function, turning
KL from a static stability knob into an exploration--stability
controller without introducing new hyperparameters beyond the
existing~$\beta$.

\paragraph{Difficulty-aware sampling and curricula in RLVR.}
A second body of work asks not how to update the policy, but
\emph{which questions to update on}.
\GRPO's default uniform sampling treats all training questions
equally, despite the fact that questions whose empirical pass rate is
near $0$ or near $1$ produce near-zero advantage and therefore weak
gradient signal.
\DAPO addresses this with \emph{binary} hard filtering, dropping
batches where every rollout is correct or every rollout is
wrong~\citep{yu2025dapo}; MathForge's \DGPO instead applies a
\emph{monotone} upweighting that favors harder problems
overall~\citep{dai2026mathforge}.
Both views miss a key structural property: the most informative
questions concentrate around \emph{moderate} difficulty
($p\!\approx\!0.5$), where the model's policy lies on its learning
frontier; binary filtering keeps too much already-mastered mass and
monotone upweighting wastes capacity on currently-intractable problems.
Our \GCS component models this frontier explicitly with a Gaussian
weighting in pass-rate space, recovering binary hard filtering and
monotone difficulty upweighting as limiting cases of a single,
continuously tunable curriculum.

Taken together, prior RLVR work optimizes either how the policy is
regularized or which problems it is trained on, but seldom both, and
typically with mechanisms that are insensitive to the model's evolving
competence.
FG-ExPO instead couples competence-conditioned KL regulation~(\AKL) with
a frontier-aware curriculum~(\GCS), so that the exploration budget is
allocated jointly across optimization strength and data distribution.

\section{Method}
\label{sec:method}

\subsection{Overview}
\label{sec:method:overview}

We build FG-ExPO on the standard RLVR training pipeline with a
verifier-based binary reward and \GRPO~\citep{shao2024deepseekmath} as
the underlying policy optimizer.
The central observation behind FG-ExPO is that two seemingly secondary
choices in \GRPO---a fixed KL coefficient and a uniform question
sampling distribution---are both insensitive to the model's evolving
competence and consistently misallocate the exploration budget.
FG-ExPO addresses this with two complementary, parameter-light components
that are conditioned on different levels of competence.
At the \emph{batch} level, Accuracy-Conditioned KL Scaling
(\AKL,~\S\ref{sec:method:akl}) replaces the fixed KL coefficient with a
nonlinearly scaled coefficient driven by the batch's mean accuracy
$\accbar$, so that the reference-model anchor is relaxed on hard
batches and tightened on easy ones.
At the \emph{question} level, Gaussian Curriculum Sampling
(\GCS,~\S\ref{sec:method:gcs}) replaces uniform sampling with a
Gaussian-shaped curriculum over an EMA-smoothed per-question pass-rate,
so that frontier-difficulty questions ($p\!\approx\!0.5$) receive the
highest sampling probability.
Combining these two components yields the unified training procedure
in \S\ref{sec:method:algo}; we present implementation details in
\S\ref{sec:method:impl}.
The remainder of this section first reviews the \GRPO objective and
the unbiased KL estimator we adopt~(\S\ref{sec:method:prelim}).

\subsection{Preliminaries}
\label{sec:method:prelim}
\GRPO eliminates the value critic of PPO~\citep{schulman2017ppo} by
estimating advantages via group-relative reward normalization.
Let $\pi_\theta$ denote the current policy and $\pi_{\mathrm{ref}}$ the
frozen reference policy.
For each training question $q$, \GRPO samples a group of $G$ rollouts
$\{o^{(g)}\}_{g=1}^{G}\sim \pi_{\theta_{\text{old}}}(\cdot\mid q)$ and
scores them with a verifier-based reward $R(o^{(g)},q)\in\{0,1\}$.
The group-relative advantage of the $g$-th rollout is
\begin{equation}
\Ahat^{(g)}
=\frac{R(o^{(g)},q)-\mu_q}{\sigma_q+\epsilon},
\quad
\mu_q=\frac{1}{G}\sum_{g=1}^{G}R(o^{(g)},q),
\quad
\sigma_q^2=\frac{1}{G}\sum_{g=1}^{G}\bigl(R(o^{(g)},q)-\mu_q\bigr)^2.
\label{eq:grpo-adv}
\end{equation}
\GRPO then optimizes the token-level clipped importance-sampled
surrogate
\begin{equation}
\mathcal{J}_{\text{clip}}(\theta)
=\mathbb{E}_{q,\,g,\,t}\!\left[
\min\!\bigl(
r_{g,t}(\theta)\,\Ahat^{(g)},\;
\mathrm{clip}\!\bigl(r_{g,t}(\theta),1{-}\varepsilon,1{+}\varepsilon\bigr)\,\Ahat^{(g)}
\bigr)
\right],
\label{eq:grpo-clip}
\end{equation}
where
$r_{g,t}(\theta)
=\pi_\theta(o^{(g)}_t\mid q,o^{(g)}_{<t})/\pi_{\theta_{\text{old}}}(o^{(g)}_t\mid q,o^{(g)}_{<t})$
is the token-level importance ratio and $\varepsilon$ is the clipping
threshold.

To prevent the optimized policy from drifting too far from
$\pi_{\mathrm{ref}}$, \GRPO adds a per-token KL penalty.
Following DeepSeekMath~\citep{shao2024deepseekmath}, we adopt the
unbiased K3 estimator
\begin{equation}
\widehat{\KLdiv}\!\bigl(\pi_\theta\,\|\,\pi_{\mathrm{ref}}\bigr)
=\frac{\pi_{\mathrm{ref}}(o_t\mid q,o_{<t})}{\pi_\theta(o_t\mid q,o_{<t})}
-\log\!\frac{\pi_{\mathrm{ref}}(o_t\mid q,o_{<t})}{\pi_\theta(o_t\mid q,o_{<t})}-1
\;\geq\;0,
\label{eq:k3}
\end{equation}
which is non-negative, has zero bias under $o\sim\pi_\theta$, and
exhibits lower variance than the naive log-ratio estimator.
Combining \eqref{eq:grpo-clip} and \eqref{eq:k3}, the standard \GRPO
objective is
$\mathcal{J}_{\text{GRPO}}(\theta)=\mathcal{J}_{\text{clip}}(\theta)
-\beta\,\mathbb{E}_{q,g,t}[\widehat{\KLdiv}]$
with a \emph{fixed} coefficient $\beta$.
FG-ExPO retains \eqref{eq:grpo-adv}--\eqref{eq:k3} unchanged and modifies
only how $\beta$ is scaled (\S\ref{sec:method:akl}) and how questions
$q$ are sampled (\S\ref{sec:method:gcs}).

\subsection{Accuracy-Conditioned KL Scaling (\AKL)}
\label{sec:method:akl}
A fixed KL coefficient $\beta$ collapses two qualitatively different
training regimes into one.
When the model fails most rollouts in a batch, $\pi_{\mathrm{ref}}$
is not a strong solution and a large $\beta$ unnecessarily anchors
$\pi_\theta$ to a weak prior, suppressing the very exploration the
batch demands; conversely, when the model succeeds, $\pi_{\mathrm{ref}}$
is closer to a useful prior and $\beta$ should be \emph{larger} to
suppress drift and preserve already-learned competence.
A single static $\beta$ cannot satisfy both regimes simultaneously,
which motivates conditioning the coefficient on the model's competence
on the current batch.

We define the batch-level accuracy as
$\accbar=\tfrac{1}{N}\sum_{i=1}^{N}\mathrm{acc}_i\in[0,1]$, the mean
verifier score over the $N$ rollouts in the current batch
($\mathrm{acc}_i\in\{0,1\}$).
\AKL replaces the fixed coefficient $\beta$ with a competence-dependent
coefficient $\beta_{\text{eff}}(\accbar)$ obtained from a smooth
nonlinear scaling function $\rho:[0,1]\rightarrow\mathbb{R}_{+}$:
\begin{equation}
\beta_{\text{eff}}(\accbar)=\beta\cdot\rho(\accbar),
\label{eq:akl}
\end{equation}
where $\rho$ is required to be (i) monotone increasing in $\accbar$
and (ii) bounded above and below by strictly positive constants
$0<\rho_{\min}\leq\rho(\cdot)\leq\rho_{\max}<\infty$.
The exact form of $\rho$ used in our experiments is given in
\eqref{eq:akl-instantiation} and instantiated in \S\ref{sec:exp:setup};
the analysis below relies only on these two structural properties.
The corresponding FG-ExPO loss is then
\begin{equation}
\mathcal{J}_{\text{FG-ExPO-KL}}(\theta)
=\mathcal{J}_{\text{clip}}(\theta)
-\beta_{\text{eff}}(\accbar)\cdot
\mathbb{E}_{q,g,t}\!\bigl[\widehat{\KLdiv}\bigr].
\label{eq:akl-loss}
\end{equation}

The two structural properties of $\rho$ jointly translate into the
desired training behavior.
Monotone competence-coupling means that harder batches receive a
smaller anchor while easier batches receive a larger one, directly
inverting the failure mode described above and producing a
self-balancing exploration--stability trade-off whose strength tracks
the policy's current competence rather than a manually picked
schedule.
Two-sided boundedness of $\rho$ further ensures that
$\beta_{\text{eff}}$ remains within the same order of magnitude as
$\beta$ throughout training, preventing both KL collapse on hard
batches and KL explosion on easy ones, and avoiding the instability
of the $\beta\!=\!0$ regime used by prior work~\citep{yu2025dapo}.
By design, \AKL preserves the original \GRPO loss structure and
modifies only the scalar multiplier on KL, so the standard \GRPO
objective is recovered exactly when $\rho\!\equiv\!1$ (and, more
generally, any constant $\rho\!\equiv\!c$ is equivalent to running
\GRPO with a rescaled coefficient $c\beta$).

\subsection{Gaussian Curriculum Sampling (\GCS)}
\label{sec:method:gcs}
\GRPO samples training questions uniformly over the dataset, but the
information content of a rollout is highly non-uniform.
For a question $q$ that the policy solves almost always
($p_q\!\to\!1$) or almost never ($p_q\!\to\!0$), the rewards
$R(o^{(g)},q)$ in \eqref{eq:grpo-adv} are nearly constant across
the group, the within-group standard deviation $\sigma_q$ collapses,
and the resulting advantages and gradients vanish.
The most informative gradients come instead from \emph{frontier}
questions where $p_q\!\approx\!0.5$, which lie at the edge of the
policy's current competence.
\DAPO addresses this only with binary $p\!\in\!\{0,1\}$
filtering~\citep{yu2025dapo} and MathForge's \DGPO with monotone
hard-up\-weighting~\citep{dai2026mathforge}; both miss the
bell-shaped structure of informativeness around $p\!=\!0.5$ and
either preserve too much already-mastered mass or waste capacity on
currently-intractable problems.

To exploit this bell-shaped structure, we maintain a smoothed
per-question pass-rate $\tilde p_q\in[0,1]$ for every question $q$ in
the training set, and use it to drive a Gaussian curriculum over the
sampling distribution.
Whenever $q$ is sampled at training step $t$ and produces an
empirical pass-rate
$p_q^{(t)}=\tfrac{1}{G}\sum_{g=1}^{G}R(o^{(g)},q)$ from its $G$
rollouts, we update
\begin{equation}
\tilde p_q^{(t)}
=\alpha\,\tilde p_q^{(t-1)}+(1-\alpha)\,p_q^{(t)},
\qquad \alpha\in[0,1),
\label{eq:ema}
\end{equation}
which damps the per-group sampling noise while still tracking the
policy's evolving competence on $q$; questions not sampled at step
$t$ retain their previous estimate.
We then map $\tilde p_q$ to a sampling weight via a Gaussian density
centered at the frontier $\mu\!=\!0.5$ and normalize over the
dataset:
\begin{equation}
w_q^{(t)}
=\exp\!\left(-\frac{(\tilde p_q^{(t)}-0.5)^2}{2\sigma^2}\right),
\qquad
\Pr\!\bigl[q\text{ sampled at step }t+1\bigr]
=\frac{w_q^{(t)}}{\sum_{q'}w_{q'}^{(t)}},
\label{eq:gcs}
\end{equation}
where $\sigma^2$ is a single curriculum-sharpness hyperparameter.
At the start of training we have no rollouts and initialize
$\tilde p_q^{(0)}\!=\!0.5$ for all $q$, which makes \eqref{eq:gcs}
recover uniform sampling until \eqref{eq:ema} populates the table.

This formulation has three structural advantages over prior
difficulty schedulers.
First, it is frontier-centered rather than extreme-centered: the
maximum of $w_q$ occurs at $\tilde p_q\!=\!0.5$, where the
group-relative advantages of \eqref{eq:grpo-adv} attain their largest
variance and the gradient signal is therefore strongest.
Second, a single $\sigma^2$ continuously interpolates between the two
extremes used by prior work---as $\sigma\!\to\!0$ the weights
degenerate into a sharp window around the frontier and recover an
aggressive analog of \DAPO's $p\!\in\!\{0,1\}$
filter~\citep{yu2025dapo}, while as $\sigma\!\to\!\infty$ the weights
become uniform and recover \GRPO's default sampler---so binary
filtering and uniform sampling are both limiting cases of \GCS rather
than separate design alternatives.
Third, the EMA in \eqref{eq:ema} prevents a single noisy rollout
group from abruptly moving $q$ in or out of the frontier and avoids
curriculum oscillation.

\subsection{FG-ExPO Training Algorithm}
\label{sec:method:algo}
Combining \AKL and \GCS yields a single training loop that differs
from \GRPO only in two places: the question sampler and the KL
coefficient.
The rollout, group-relative advantage, and clipped-surrogate
computations all remain unchanged from \eqref{eq:grpo-adv}--\eqref{eq:k3},
so FG-ExPO can be implemented as a drop-in modification to any
\GRPO-style RLVR pipeline.
Algorithm~\ref{alg:fg-expo} summarizes the full procedure.

\begin{algorithm}[t]
\color{black}%
\caption{FG-ExPO: Frontier-Guided Exploration-Prioritized Policy Optimization}
\label{alg:fg-expo}
\begin{algorithmic}[1]
\Require Dataset $\mathcal{D}=\{q\}$, reference policy $\pi_{\mathrm{ref}}$,
initial policy $\pi_\theta$, group size $G$, KL coefficient $\beta$,
nonlinear KL scaling function $\rho$, clip $\varepsilon$, EMA factor
$\alpha$, curriculum width $\sigma^2$, training steps $T$.
\State Initialize $\tilde p_q^{(0)}\!\gets\!0.5$ for all $q\in\mathcal{D}$.
\For{$t=1,\dots,T$}
  \State \textbf{(GCS)} compute weights
    $w_q^{(t-1)}\!\gets\!\exp\!\bigl(-(\tilde p_q^{(t-1)}-0.5)^2/(2\sigma^2)\bigr)$
    and sample a question batch
    $\mathcal{B}_t\!\sim\!\Pr[q]\!\propto\!w_q^{(t-1)}$.
  \For{each $q\in\mathcal{B}_t$}
    \State Sample $G$ rollouts $\{o^{(g)}\}\sim\pi_{\theta_{\text{old}}}(\cdot\mid q)$
      and score $R(o^{(g)},q)$ via the verifier.
    \State Compute group-relative advantages $\Ahat^{(g)}$ via
      Eq.~\eqref{eq:grpo-adv}.
    \State Update
      $\tilde p_q^{(t)}\!\gets\!\alpha\,\tilde p_q^{(t-1)}+(1-\alpha)\,
      \tfrac{1}{G}\sum_g R(o^{(g)},q)$.
  \EndFor
  \State Compute batch mean accuracy
    $\accbar\!\gets\!\tfrac{1}{|\mathcal{B}_t|G}\sum_{q,g}R(o^{(g)},q)$.
  \State \textbf{(AKL)} set
    $\beta_{\text{eff}}\!\gets\!\beta\cdot\rho(\accbar)$.
  \State Compute clipped surrogate $\mathcal{J}_{\text{clip}}(\theta)$
    via Eq.~\eqref{eq:grpo-clip} and the K3 estimator
    $\widehat{\KLdiv}$ via Eq.~\eqref{eq:k3}.
  \State Update $\theta$ by gradient ascent on
    $\mathcal{J}_{\text{clip}}(\theta)-\beta_{\text{eff}}\,
     \mathbb{E}\bigl[\widehat{\KLdiv}\bigr]$.
\EndFor
\State \Return $\pi_\theta$.
\end{algorithmic}
\end{algorithm}

\subsection{Implementation Details}
\label{sec:method:impl}
FG-ExPO inherits all GRPO-side hyperparameters from
DeepSeekMath~\citep{shao2024deepseekmath} (clip threshold
$\varepsilon$, group size $G$, learning rate, optimizer) without
modification, so the only FG-ExPO-specific knobs are the EMA factor
$\alpha$, the curriculum width $\sigma^2$, and the choice of nonlinear
scaling function $\rho$ used in
Eqs.~\eqref{eq:ema}--\eqref{eq:gcs}.
The smoothed pass-rate $\tilde p_q$ is refreshed once per rollout
step and the sampling distribution is recomputed on the fly, so \GCS
adds only $O(|\mathcal{D}|)$ per-step bookkeeping and no extra
forward or backward passes.
Concrete values of $\beta$, $\alpha$, $\sigma^2$ and the instantiation
of $\rho$ used in our experiments are listed in
\S\ref{sec:exp:setup}.

\section{Experiments}
\label{sec:experiments}

\subsection{Experimental Setup}
\label{sec:exp:setup}

We train FG-ExPO and the \GRPO baseline on the DAPO-17K mathematical
reasoning training set introduced by \DAPO~\citep{yu2025dapo}, and
evaluate on six competition-grade benchmarks: AIME~2024, AIME~2025,
MATH-500, Minerva, OlympiadBench, and AMC.
To assess robustness across model families and scales, we use two
base models---DeepSeek-R1-Distill-Qwen-1.5B~\citep{deepseekai2025r1distill}
and Qwen3-8B-Base~\citep{yang2025qwen3}---and run identical training
pipelines for the two methods, differing only in FG-ExPO's two components
(\AKL and \GCS); the reference policy $\pi_{\mathrm{ref}}$ is the
corresponding base model and is kept frozen throughout training.

Unless otherwise stated, we instantiate the nonlinear \AKL scaling
function as
\begin{equation}
\rho(x)=\frac{\tanh(x)+1}{2},
\qquad x\in[0,1],
\label{eq:akl-instantiation}
\end{equation}
which maps the batch accuracy to a bounded coefficient range
$\rho(x)\in[0.5,\,0.881]$, halving the effective KL strength on
fully-failing batches and tightening it to roughly $0.88\beta$ on
fully-succeeding ones.
For \GCS we use a Gaussian curriculum with mean $\mu=0.5$ and
standard deviation $\sigma=0.35$ (so the curriculum width in
Eq.~\eqref{eq:gcs} is $\sigma^2\!\approx\!0.1225$), and refresh the
EMA-smoothed pass-rates $\tilde p_q$ via Eq.~\eqref{eq:ema} once per
training step with EMA factor $\alpha=0.9$.

For training we use group size $G\!=\!8$ rollouts per question,
question batch size $|\mathcal{B}|\!=\!256$, base KL coefficient
$\beta\!=\!0.02$ (modulated by \AKL into $\beta_{\text{eff}}$),
maximum prompt length $1024$, and maximum response length $3072$.
The KL term uses the unbiased K3 estimator of Eq.~\eqref{eq:k3} and
is applied inside the loss rather than the reward, following the
established convention in
RLVR-style policy optimization~\citep{shao2024deepseekmath,yu2025dapo,liu2025drgrpo}.
We train for $30$ epochs on a single node with $8$ GPUs, using the
verl framework~\citep{sheng2024verl} for distributed RL training and
vLLM with tensor-parallel size $2$ for rollout generation.
All remaining optimizer and scheduling hyperparameters follow the
DAPO-17K reference recipe~\citep{yu2025dapo} for both methods,
isolating the contribution of \AKL and \GCS.

For every benchmark we draw $32$ independent completions per question
with sampling temperature $0.6$ and the same maximum response length
used at training, and report two metrics from the \emph{same}
$32$-sample inference budget.
\textbf{pass@1} is the unbiased single-sample success probability,
estimated as the mean verifier reward over the $32$ completions
(equivalently, the average accuracy of independent
attempts);
\textbf{pass@32} is the empirical probability that at least one of
the $32$ completions is correct.
pass@1 thus captures the policy's expected performance under a
\emph{single} attempt, while pass@32 captures the breadth of correct
reasoning paths the policy can produce when test-time compute is
allowed to grow---a regime that has been shown to dramatically
expand model capability in recent test-time scaling
work~\citep{brown2024largescale,snell2024scaling}.
The gap between the two metrics, computed from the same sample pool,
therefore quantifies the policy's \emph{effective exploration space},
which is the central quantity our method targets.

\subsection{Main Results}
\label{sec:exp:main}

\begin{table}[t]
\centering
\small
\setlength{\tabcolsep}{4pt}
\renewcommand{\arraystretch}{1.05}
\caption{Main results of \GRPO and FG-ExPO on six competition-grade
mathematical reasoning benchmarks using DeepSeek-R1-Distill-Qwen-1.5B
and Qwen3-8B-Base, reported as pass@1 and pass@32 accuracy~(\%).}
\label{tab:main}
\begin{tabular}{l|cccccc|c}
\toprule
\textbf{Method} & \textbf{AIME 2024} & \textbf{AIME 2025} & \textbf{MATH-500} & \textbf{Minerva} & \textbf{Olympiad} & \textbf{AMC} & \textbf{Avg} \\
\midrule
\multicolumn{8}{c}{\textbf{pass@1}} \\
\cmidrule(lr){1-8}
\multicolumn{8}{l}{\itshape DeepSeek-R1-Distill-Qwen-1.5B} \\
\quad Base               & 20.73 & 18.54 & 73.44 & 21.48 & 22.10 & 52.02 & 34.72 \\
\quad + \GRPO            & 27.40 & 22.60 & 82.88 & 26.86 & \textbf{30.23} & 68.45 & 43.07 \\
\quad + \textbf{FG-ExPO}    & \textbf{29.06} & \textbf{23.12} & \textbf{84.03} & \textbf{26.88} & 29.33 & \textbf{68.75} & \textbf{43.53} \\
\quad $\Delta$           & \posdelta{+1.66} & \posdelta{+0.52} & \posdelta{+1.15} & \posdelta{+0.02} & \negdelta{$-$0.90} & \posdelta{+0.30} & \posdelta{\textbf{+0.46}} \\
\addlinespace
\multicolumn{8}{l}{\itshape Qwen3-8B-Base} \\
\quad + \GRPO            & 54.58 & 40.00 & 91.78 & 38.72 & \textbf{46.38} & 82.47 & 58.99 \\
\quad + \textbf{FG-ExPO}    & \textbf{57.29} & \textbf{43.33} & \textbf{92.16} & \textbf{39.11} & 46.25 & \textbf{82.70} & \textbf{60.14} \\
\quad $\Delta$           & \posdelta{+2.71} & \posdelta{+3.33} & \posdelta{+0.38} & \posdelta{+0.39} & \negdelta{$-$0.13} & \posdelta{+0.23} & \posdelta{\textbf{+1.15}} \\
\midrule
\multicolumn{8}{c}{\textbf{pass@32}} \\
\cmidrule(lr){1-8}
\multicolumn{8}{l}{\itshape DeepSeek-R1-Distill-Qwen-1.5B} \\
\quad Base               & 63.33 & 36.67 & 93.60 & 46.69 & 44.00 & 85.37 & 61.61 \\
\quad + \GRPO            & 73.33 & 56.67 & 95.00 & \textbf{48.53} & 54.00 & \textbf{92.68} & 70.04 \\
\quad + \textbf{FG-ExPO}    & \textbf{83.33} & \textbf{60.00} & \textbf{96.00} & 47.06 & \textbf{55.33} & 91.46 & \textbf{72.20} \\
\quad $\Delta$           & \posdelta{+10.00} & \posdelta{+3.33} & \posdelta{+1.00} & \negdelta{$-$1.47} & \posdelta{+1.33} & \negdelta{$-$1.22} & \posdelta{\textbf{+2.16}} \\
\addlinespace
\multicolumn{8}{l}{\itshape Qwen3-8B-Base} \\
\quad + \GRPO            & \textbf{80.00} & 63.33 & \textbf{96.80} & 49.63 & 59.33 & 95.12 & 74.04 \\
\quad + \textbf{FG-ExPO}    & \textbf{80.00} & \textbf{76.67} & 96.80 & \textbf{50.37} & \textbf{60.00} & \textbf{96.34} & \textbf{76.70} \\
\quad $\Delta$           & 0.00 & \posdelta{+13.34} & 0.00 & \posdelta{+0.74} & \posdelta{+0.67} & \posdelta{+1.22} & \posdelta{\textbf{+2.66}} \\
\bottomrule
\end{tabular}
\end{table}

FG-ExPO consistently outperforms \GRPO across both scales and both
inference budgets.
On the 1.5B backbone, FG-ExPO improves the average accuracy by
$+0.46$ on pass@1 and $+2.16$ on pass@32; on the 8B backbone, the
gains are $+1.15$ on pass@1 and $+2.66$ on pass@32.
The per-benchmark margins are largest on the most challenging
competition tasks: on AIME~2024 and AIME~2025, FG-ExPO improves pass@1
by $+1.66$/$+0.52$ on the 1.5B model and $+2.71$/$+3.33$ on the 8B
model, and improves pass@32 by $+10.00$/$+3.33$ on the 1.5B model
and $0.00$/$+13.34$ on the 8B model.
On more saturated benchmarks (MATH-500 and AMC), gains are smaller
and occasionally near-tied---e.g., the 8B pass@32 score on
AIME~2024 already reaches $80.00$ for both methods---which is
consistent with FG-ExPO targeting the frontier of the policy's current
competence rather than benchmarks the model already saturates.
A small number of cells show minor regressions
(e.g., 1.5B pass@32 on Minerva: $-1.47$;
1.5B pass@1 on OlympiadBench: $-0.90$), but these are dominated by
gains on the harder benchmarks, and the average across all six
benchmarks is strictly improved in every model$\times$metric setting.

\subsection{Exploration Effect on Mathematical Reasoning RLVR}
\label{sec:exp:explore}

A central prediction of FG-ExPO is that reallocating exploration
budget---both via competence-conditioned KL and frontier-centered
sampling---should expand the diversity of correct reasoning paths
the policy can discover on math RLVR, not merely refine its single
most likely solution.
Because pass@1 and pass@32 in Table~\ref{tab:main} are computed from
the same $32$-sample budget, their gap is a direct empirical
measurement of how much benefit the policy can extract from
additional inference-time compute, in the spirit of recent
test-time scaling
results~\citep{brown2024largescale,snell2024scaling}; a method that
genuinely expands the policy's reasoning frontier should therefore
yield \emph{disproportionately larger} gains on pass@32 than on
pass@1.
Our results match this prediction on both scales: on the 1.5B model
FG-ExPO improves the benchmark average by $+0.46$ on pass@1 but $+2.16$
on pass@32, a roughly $4.7\!\times$ ratio; on the 8B model the same
comparison yields $+1.15$ versus $+2.66$, a $2.3\!\times$ ratio.
At the per-benchmark level, the largest pass@32 gains coincide
exactly with the hardest competition tasks---$+10.00$ on
AIME~2024 (1.5B) and $+13.34$ on AIME~2025 (8B)---i.e., the
regime in which the policy still has the most room left to expand
its set of correct reasoning paths and in which \AKL has the most
competence-conditioned slack to relax the reference-model anchor.
Together, these results indicate that FG-ExPO does not merely sharpen
the policy's most likely answer on math RLVR; it broadens the set of
correct reasoning paths the policy can produce within a fixed inference
budget, addressing precisely the failure mode that fixed-coefficient KL
and uniform question sampling jointly induce in standard \GRPO.

\subsection{Ablation Study}
\label{sec:exp:ablation}

FG-ExPO introduces two complementary components, \AKL on the optimization
side and \GCS on the data side, that target the same exploration
miscalibration through different levers.
To verify that each component contributes to the overall gain rather
than one masking the other's failure, we ablate the two components
individually on the strongest backbone (Qwen3-8B-Base) and report
pass@32 on all six benchmarks.
We adopt pass@32 as the primary metric here because, as established in
\S\ref{sec:exp:explore}, it most directly measures the reasoning-path
diversity that FG-ExPO is designed to expand.
Table~\ref{tab:ablation} summarizes the four configurations: the
\GRPO baseline, FG-ExPO with one component disabled at a time, and the
full FG-ExPO model.

\begin{table}[t]
\centering
\small
\setlength{\tabcolsep}{4pt}
\renewcommand{\arraystretch}{1.05}
\caption{Ablation of \AKL and \GCS on Qwen3-8B-Base, reported as
pass@32 accuracy~(\%) on six competition-grade benchmarks.}
\label{tab:ablation}
\begin{tabular}{l|ccccccc|c}
\toprule
\textbf{Method} & \textbf{AIME 2024} & \textbf{AIME 2025} & \textbf{MATH-500} & \textbf{Minerva} & \textbf{Olympiad} & \textbf{AMC} & \textbf{Avg} & $\boldsymbol{\Delta_{\GRPO}}$ \\
\midrule
\GRPO                 & 80.00             & 63.33             & 96.80             & 49.63             & 59.33             & 95.12             & 74.04             & ---                \\
FG-ExPO (\emph{w/o \GCS})  & 80.00             & 73.33             & 96.80             & 48.90             & 58.00             & 93.90             & 75.16             & \posdelta{+1.12}              \\
FG-ExPO (\emph{w/o \AKL})  & \textbf{83.33}    & \underline{73.33} & \textbf{97.00}    & \underline{49.63} & \underline{59.33} & \underline{95.12} & \underline{76.29} & \posdelta{+2.25}  \\
FG-ExPO (full)           & \underline{80.00} & \textbf{76.67}    & \underline{96.80} & \textbf{50.37}    & \textbf{60.00}    & \textbf{96.34}    & \textbf{76.70}    & \posdelta{\textbf{+2.66}}     \\
\bottomrule
\end{tabular}
\end{table}

Both components are individually beneficial.
Disabling \GCS but retaining \AKL improves the benchmark average from
$74.04$ to $75.16$~($+1.12$), with the gain almost entirely
concentrated on AIME~2025 ($63.33\!\rightarrow\!73.33$); this matches
our analysis in \S\ref{sec:method:akl} that competence-conditioned KL
matters most when the policy is far from the reference and most
batches are failing, exactly the regime that AIME~2025 induces on a
freshly-trained 8B model.
Conversely, disabling \AKL but retaining \GCS improves the average to
$76.29$~($+2.25$), and is the only configuration that lifts AIME~2024
above the \GRPO baseline ($80.00\!\rightarrow\!83.33$).
This component-level pattern is consistent with the design split:
\AKL controls how aggressively the policy may deviate per update on a
given batch, whereas \GCS controls which questions are presented in
the first place.
Removing either component therefore degrades a different axis of
exploration, and neither alone closes the gap to the full model.

The full FG-ExPO configuration outperforms both single-component
variants, achieving the best benchmark average ($76.70$, $+2.66$ over
\GRPO) and the strongest AIME~2025 score ($76.67$, $+13.34$ over
\GRPO).
The two components are clearly complementary rather than redundant:
neither \GCS alone nor \AKL alone reaches the full model's average,
and the full model attains the highest score on four of the six
benchmarks (AIME~2025, Minerva, OlympiadBench, and AMC), while the
\emph{w/o \AKL} variant takes the lead on AIME~2024 ($83.33$ vs.
$80.00$) and MATH-500 ($97.00$ vs.\ $96.80$).
We attribute this complementarity to the fact that \GCS supplies the
\emph{frontier-difficulty batches} on which competence-conditioned KL
has the most slack to relax---i.e., \GCS controls the input
distribution to \AKL---so the joint allocation of exploration budget
across data and optimization is strictly broader in capacity than
either allocation alone.
The smaller margin of FG-ExPO~(full) over the \emph{w/o \AKL} variant
($+0.41$ on average) compared to its margin over \emph{w/o \GCS}
($+1.54$) further suggests that, on this 8B backbone, \GCS is the
more decisive of the two components on average, while \AKL provides
additional headroom precisely on the hardest benchmark
(most visibly the $63.33\!\to\!76.67$ jump on AIME~2025).

Two takeaways emerge.
First, the largest per-benchmark gains across the ablation occur on
AIME~2024 and AIME~2025, the two most challenging competition
benchmarks at this scale and the regime in which the policy has the
most room left to expand its set of correct reasoning paths;
nearly-saturated benchmarks such as MATH-500 (\GRPO already at
$96.80$) move by at most $0.20$ percentage points across
configurations---confirming that FG-ExPO's gains come from reallocating
exploration toward the policy's actual frontier rather than from
scale-invariant regularization tweaks.
Second, the per-component deltas (\AKL alone $+1.12$, \GCS alone
$+2.25$) sub-additively combine into the full-model delta ($+2.66$),
which is the expected signature of two interacting mechanisms that
target the same scarce resource (exploration budget) from different
directions.
We therefore recommend deploying \AKL and \GCS jointly; using either
alone retains most of the implementation simplicity but only a
fraction of the gain.

\section{Conclusion}
\label{sec:conclusion}

We revisited two design choices that \GRPO and most of its descendants
inherit unchanged---a fixed KL coefficient and a uniform question
sampler---and identified them as a single, unifying source of
exploration miscalibration in math RLVR.
We addressed this with FG-ExPO, a unified extension of \GRPO that
couples \emph{Accuracy-Conditioned KL Scaling}~(\AKL), a smooth
nonlinear function of the batch's mean accuracy that relaxes the
reference-model anchor on hard batches and tightens it on easy ones,
with \emph{Gaussian Curriculum Sampling}~(\GCS), a Gaussian-shaped
weighting in pass-rate space that concentrates training on
frontier-difficulty questions ($p\!\approx\!0.5$).

Across two base models of distinct families and scales
(DeepSeek-R1-Distill-Qwen-1.5B and Qwen3-8B-Base) and six
competition-grade benchmarks, FG-ExPO consistently outperforms \GRPO,
with gains that are largest on the hardest evaluations and on the
broader pass@32 metric---most notably a $+13.34$ absolute improvement
on AIME~2025 pass@32 and $+2.66$ average pass@32 gain on the 8B
backbone.
The disproportionately larger gains on pass@32 than on pass@1, paired
with the component ablation, support our central claim that FG-ExPO
expands the policy's effective exploration space rather than merely
sharpening its single most likely answer.
Because \AKL and \GCS together touch only the KL coefficient and the
question sampler, FG-ExPO is a drop-in modification to any
\GRPO-style RLVR pipeline and adds no extra forward or backward
passes per training step.

Our evaluation is currently bounded to mathematical reasoning RLVR
with binary verifier rewards on the DAPO-17K corpus and two model
backbones, and the specific shapes of \AKL's nonlinear scaling
function $\rho$ and \GCS's Gaussian curriculum are tuned for this
regime; whether the same shapes transfer to denser or noisier reward
signals---e.g., partial-credit reward shaping, code execution traces,
or multi-turn agentic tasks---remains an open question.
Two natural next steps follow.
First, extending FG-ExPO's competence-conditioning principle to
\emph{token}- or \emph{trajectory}-level granularity, so that KL
strength and curriculum weighting adapt to within-rollout
informativeness rather than only batch-level mean accuracy.
Second, integrating FG-ExPO with orthogonal RLVR improvements---such as
adaptive group sizes, adaptive clipping schedules, or
exploration-aware reward shaping---to study how far the joint
allocation of exploration budget across data, optimization, and
reward design can be pushed under a fixed compute envelope.

\bibliographystyle{plainnat}
\bibliography{references}

\end{document}